\documentclass[10pt,twocolumn,letterpaper]{article}

\usepackage{wacv}
\usepackage{times}
\usepackage{epsfig}
\usepackage{graphicx}
\usepackage{amsmath}
\usepackage{amssymb}

\usepackage{float}
\usepackage[utf8]{inputenc}
\usepackage{xpatch}

\usepackage{algorithm}
\usepackage{algcompatible}
\usepackage[noend]{algpseudocode}
\usepackage{booktabs}

\usepackage{xcolor}
\usepackage[bookmarks=false]{hyperref}
\usepackage{url}

\makeatletter
\xpatchcmd{\algorithmic}
  {\ALG@tlm\z@}{\leftmargin\z@\ALG@tlm\z@}
  {}{}
\makeatother

\wacvfinalcopy 

\def\wacvPaperID{314} 

\ifwacvfinal\fi
\begin{document}
\title{TailorGAN: Making User-Defined Fashion Designs}

\author{
Lele Chen$^\dagger$ \hspace{0.5cm} Justin Tian$^\dagger$ \hspace{0.5cm} Guo Li\\
University of Rochester\\
{\tt\small lchen63@ur.rochester.edu, jtian@pas.rochester.edu, gli27@ur.rochester.edu}
\and 
Cheng-Haw Wu \hspace{0.5cm} Erh-Kan King \hspace{0.5cm} Kuan-Ting Chen \hspace{0.5cm} Shao-Hang Hsieh\\
Viscovery\\
{\tt\small \{jason.wu,jeff.king,dannie.chen,shao\}@viscovery.com}
\and 
Chenliang Xu\\
University of Rochester\\
{\tt\small chenliang.xu@rochester.edu}\\
{\tt\small $^\dagger$equal contribution}
}

\maketitle
\ifwacvfinal\thispagestyle{empty}\fi


\begin{abstract}
Attribute editing has become an important and emerging topic of computer vision. In this paper, we consider a task: given a reference garment image A and another image B with target attribute (collar/sleeve), generate a photo-realistic image which combines the texture from reference A and the new attribute from reference B. The highly convoluted attributes and the lack of paired data are the main challenges to the task. To overcome those limitations, we propose a novel self-supervised model to synthesize garment images with disentangled attributes (\eg, collar and sleeves) without paired data. Our method consists of a reconstruction learning step and an adversarial learning step. The model learns texture and location information through reconstruction learning. And, the model's capability is generalized to achieve single-attribute manipulation by adversarial learning. Meanwhile, we compose a new dataset, named GarmentSet, with annotation of landmarks of collars and sleeves on clean garment images. Extensive experiments on this dataset and real-world samples demonstrate that our method can synthesize much better results than the state-of-the-art methods in both quantitative and qualitative comparisons.

\end{abstract}

\section{Introduction}
\label{sec:intro}
\begin{figure}[t]
\includegraphics[width=0.75\linewidth]{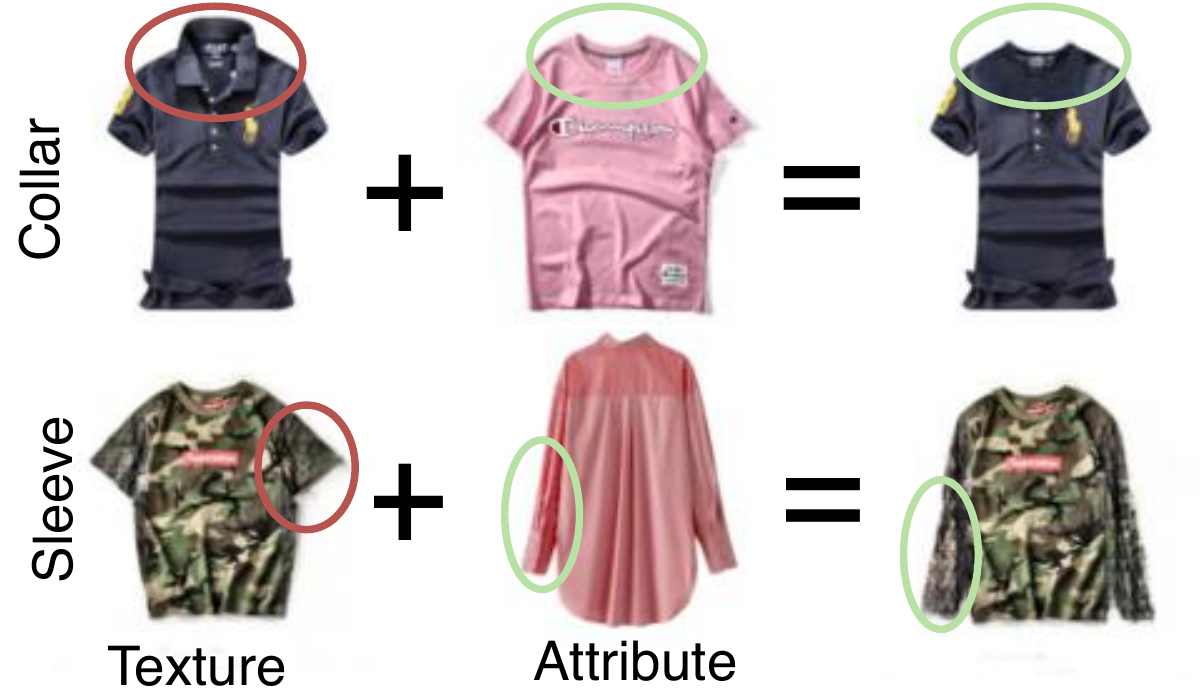}
\centering
\caption{Illustration of our task and results. 
Given a reference fashion garment (on the left) and a desired design attribute (in the middle), we aim to generate a new garment that seamlessly integrates the desired design attribute to the reference image. The first row shows collar-editing, and the second row shows sleeve-editing. Our results are shown on the right.}
\label{teaser}
\vspace{-5mm}
\end{figure}

Deep generative techniques~\cite{GAN,kingma2013auto} have led to highly successful image/video generation, some focusing on style transfer~\cite{cycleGAN}, and others on the synthesis of desired conditions~\cite{cGAN1,cGAN2}. In this paper, we propose a novel schema to disentangle attributes and synthesize high-quality images with the desired attribute while keeping other attributes unchanged. Concretely, we focus on the problem of fashion image attribute manipulation to demonstrate the capability of our method. By our method, users can switch a specific part of garments to the wanted designs (\eg, round collar to V-collar). The objective is to synthesize a photo-realistic new fashion image by combining different parts seamlessly together. Potential applications of such system range from item retrieval to professional fashion design assistant. Fig.~\ref{teaser} shows our task and results. 

Recently, Generative Adversarial Networks (GANs)~\cite{GAN, cGAN1, cGAN2}, image-to-image translation~\cite{p2p}, and CycleGAN~\cite{cycleGAN} have proved their effectiveness in generating photo-realistic images. However, image syntheses using such models usually involve highly-entangled attributes and may fail in editing target-specific attributes/objects~\cite{Face01, Face02, cycleGAN}. Meanwhile, the large variance in the garment textures causes additional problems. It is impossible to build a dataset that is large enough to approximate the distribution of all garment texture and design combinations, which serve as paired samples to train models such as~\cite{cGAN2, p2p}. Novel learning paradigm is expected to overcome this difficulty. 

To solve these challenges, we propose a novel self-supervised image generative model, TailorGAN, that makes disentangled attribute manipulations. The model exploits the latent space expression of input images without paired training images. Image edge maps are used to isolate the structures from textures. By using edge maps, we achieve good results with only a small amount of data. It is also robust to a small amount of geometric misalignment between the reference and design attribute images. The model is further generalized in a GAN framework to achieve single-attribute manipulation using random fashion inputs. Besides, an attribute-aware discriminator is weaved into the model to guide the image editing. This attribute-aware discriminator helps in making high-quality single attribute editing and guides a better self-supervised learning process. 

Current available fashion image datasets (\eg, DeepFashion~\cite{recomm3}) mostly consist of street photos with complex backgrounds or with user's body parts presented. That extra visual information may hinder the performance of an image synthesis model. Thus, to simplify the training data and screen out noisy backgrounds, we introduce a new dataset, GarmentSet. The new dataset contains fashion images with no human user presented, and most images have single-color backgrounds. 

Contributions made in this paper can be summarized as:
\begin{itemize}
\item We propose a new task in deep learning fashion studies. Instead of generating images with text guidance, virtual try-on, or texture transferring, our task is to make new fashion designs with disentangled user-appointed attributes.
\item We exploit a novel training schema consists of reconstruction learning and adversarial learning. The self-supervised reconstruction learning guides the network to learn shape, location information from the edge map, and texture information from the RGB image. The unpaired adversarial learning gives the network generalizability to synthesize images with new disentangled attributes. Besides, we propose a novel attribute-aware discriminator, which helps high-quality attribute editing by isolating the design structures from the garment textures.
\item A new dataset, GarmentSet, is introduced to serve our attribute editing task. Unlike existing fashion datasets in which most images illustrate the user's face or body parts with complicated backgrounds, GarmentSet filters out the most redundant information and directly serves the fashion design purpose.
\end{itemize}
The rest of this paper is organized as follows: A brief review of related work is presented in Sec.~\ref{sec:related}. The details of the proposed method are described in Sec.~\ref{sec:model}. We introduce our new dataset in Sec.~\ref{sec:dataset}. Experimental details and results are presented in Sec.~\ref{sec:experiments}. In Sec.~\ref{sec:ablation}, we further conduct ablation studies to investigate the performances of our model explicitly. And finally, Sec.~\ref{sec:conclution} concludes the paper with discussions of limitations. Our code and data are available at: {\small \url{https://www.cs.rochester.edu/~cxu22/r/tailorgan/}}.

\section{Related Work}
\label{sec:related}

\noindent \textbf{Generative Adversarial Network (GAN)}~\cite{GAN} is one of the most popular deep generative models and has shown impressive results in image synthesis studies, like image editing~\cite{face, indoor} and fine-grained objects generation~\cite{cGAN2,kato2019gans}. Researchers utilize different conditions to generate images with desired properties. Existing works have explored various conditions, from category labels~\cite{cLabel}, audio~\cite{cAudio1, cAudio2}, text~\cite{cGAN2}, skeleton~\cite{ma2017pose,jetchev2017conditional,zhao2018multi,raj2018swapnet} to attributes~\cite{faceAttr}. There are a few studies that investigate the task of image translations using cGAN~\cite{cGAN1, cGAN2, cGAN3}. In the context of fashion-related applications, researchers apply cGAN in automated garment textures filling~\cite{TextureGAN}, texture transferring~\cite{stlyeTransfer}, and virtual try-on~\cite{class4, tryon} by replacing dress on a person with a new one. More related work is sequential attention GAN proposed by Cheng \etal~\cite{SeqAttnGAN}. Their model uses text as the guidance and continuously changes the fashion designs based on user's requests, but the attribute changes are highly entangled. In contrast to this work, we propose a new training algorithm that combines self-supervised reconstruction learning with adversarial learning to make disentangled attribute manipulations with user-appointed images.

\begin{figure*}[t]
\centering
\includegraphics[width=0.95\linewidth]{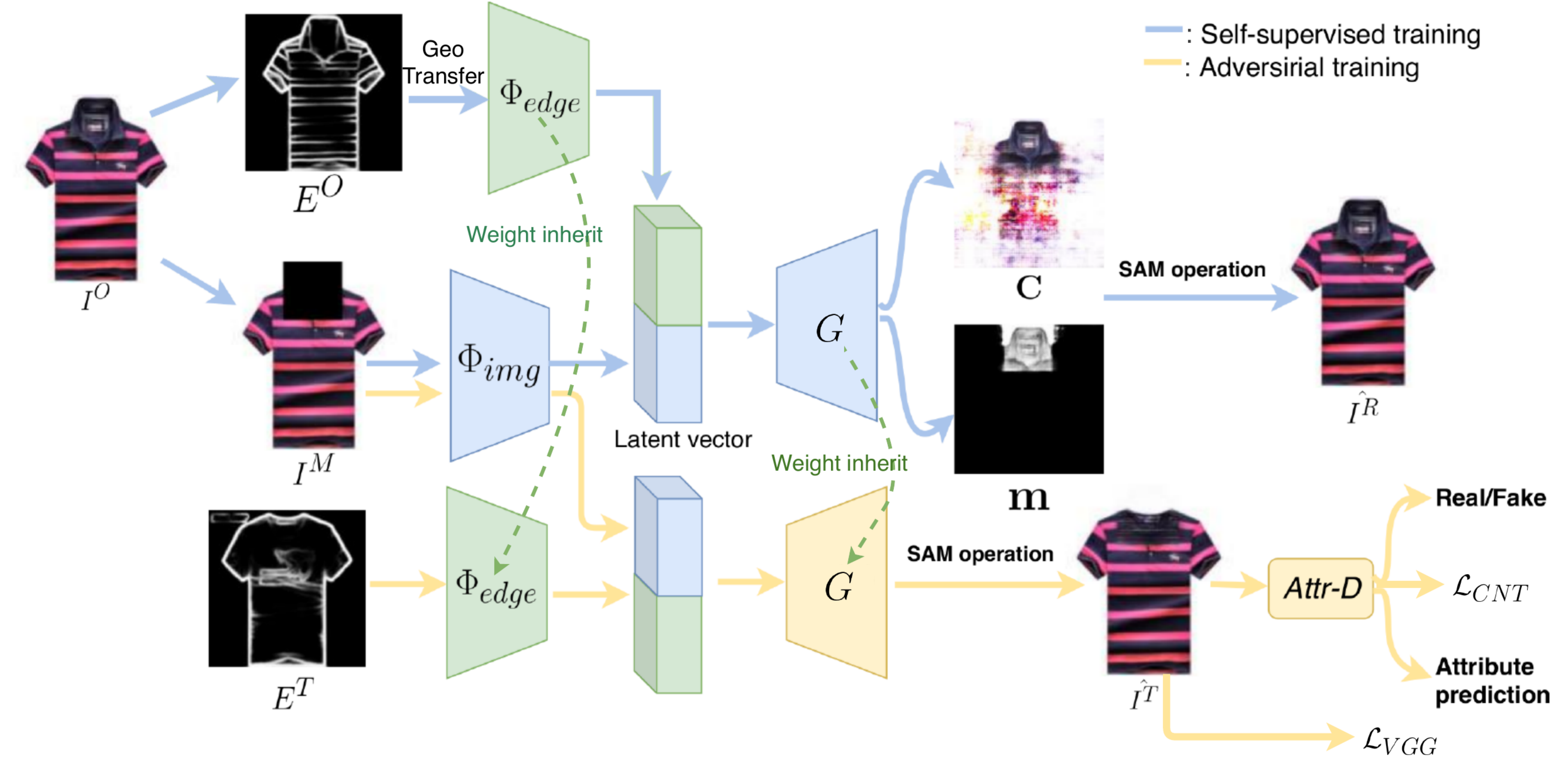}
\caption{The network architecture of TailorGAN. The Geo-Transfer indicates random rotations, translations, and scale shifts applied on the edge map. This operation can make the trained model robust to different edge maps. The weight inherit means the weights of $\Phi_{edge}$ and $G$ in adversarial step are inherited from the reconstruction step. In the reconstruction step (the upper part), we extract the edge feature (\eg location and shape of the collar) by $\Phi_{edge}$ and extract image feature (\eg overall texture and design of the garment) by $\Phi_{img}$. Then, we merge the two vectors in the latent space and pass the resultant vector through image generator $G$ to output mask and attention. In the adversarial training step (the lower part), we use $\Phi_{edge}$ to extract features from the target-attribute edge map $E^T$, which has a new collar. The attribute discriminator (Attr-$D$) outputs a real/fake score and classifies the collar type of the newly-generated image $\hat{I}^T$ yield by $G$.}
\label{network}
\end{figure*}

\noindent \textbf{Self-supervised generation} is recently introduced as a novel and effective way to train generative models without paired training data. Unpaired image-to-image translation framework such as CycleGAN~\cite{cycleGAN} removes pixel-level supervision. In CycleGAN, the unpaired image to image translation is achieved by enforcing a bi-directional translation between two domains with an adversarial penalty on the translated image in the target domain. The CycleGAN variants~\cite{cycle1,cycle2} are moving towards the direction of unsupervised learning approaches. However, CycleGAN-family models also create unexpected or even unwanted results, which are shown in the experiment section. One reason for such a phenomenon is due to the lack of straightforward knowledge of the target translation domain in the circularity training process. Inherent attributes of the source samples may be changed in a translation process. To avoid such unwanted changes, we keep an image reconstruction penalty in our image editing task. 

Attracted by the huge profit potentials in the fashion industries, deep learning methods have been conducted on fashion analysis and fashion image synthesis. Most existing researches focus on fashion trend prediction~\cite{trend}, clothing recognition with landmarks~\cite{recomm3}, clothing matching with fashion items in street photos~\cite{class1} and fashion recommendation system~\cite{recomm1,recomm2}. Different from those research lines, we focus on the fashion image synthesis task with user-appointed attribute manipulations. 

\section{Methodology}
\label{sec:model}

This section presents the implementation details of our method. There are two steps in the model training: (1) self-supervised reconstruction learning and (2) generalized attribute manipulations using adversarial learning. The first step helps the model fill correct texture for the user-appointed design pattern. The second step helps the model generate high-quality images with desired new attributes. Although we use collar translating example in Fig.~\ref{network}, we emphasize here that our model can be applied to other design patterns as shown in experiments. 

\subsection{Learning to Manipulate Designs}
\label{subsec:encoder-decoder}

\noindent \textbf{Self-Supervised Learning Step.} \quad  The motivation of formulating a self-supervised model is that it is almost impossible to collect paired training images for a fully supervised model. Using the collar editing task as an example, for each image in a fully supervised training process, one needs to collect paired images for each collar type. In these paired images, only the collar parts are different while the other attributes, like body decoration, clothing texture, etc., must stay unchanged and match with other paired images. Such data is usually unavailable. Also, the dataset size increases exponentially when multiple attribute annotations are needed for each image.

Therefore, we employ an encoder-decoder structure for the self-supervised reconstruction training step. Given a masked garment image $I^M$ (mask out the collar/sleeve part from the original garment image $I^O$) and edge map $E^{O}$, our reconstruction step reconstructs the original garment image (collar region). Intuitively, individual fashion designs may correlate with clothing texture and other design factors. For instance, light pink color is rarely used on men's garments, and leather is usually used to make jackets, etc. In our task, we want our model focus only on the design structure editing rather than the clothing texture translating, which is inherited from the reference garment image. Specifically, the self-supervised learning step is defined as:
\begin{align}
\vspace{-2mm}
\label{eq:encoder1}
\hat{I}^{R} =SAM(G( \Phi_{img}(I^M) \oplus \Phi_{edge}(E^O) ), I^M)\enspace,
\vspace{-2mm}
\end{align}
where $\Phi_{img}$ and $\Phi_{edge}$ are image encoder and edge encoder, respectively. $\Phi_{img}$ and $\Phi_{edge}$ consist of several 2D convolution layers and residual blocks. $G$ is the image decoder/generator, which consists of several 2D transpose-convolution layers. $\oplus$ is channel-wise concatenation. After encoding, we feed the concatenated latent vector to $G$ to output attention mask $\bf m$ and new pixel $\bf C$ . The SAM block (see Eq.~\ref{eq:sam}) outputs the reconstructed image $\hat{I}^R$ based on $\bf m$, $\bf C$ and $I^M$. This learning step learns how to reconstruct $I^{O}$ according to the texture feature from $I^M$ and shape feature from $E^{O}$.

Different from other methods~\cite{cAudio1,cAudio2}, we only use pixel-wise loss here to supervise the reconstruction step. Specifically, the loss function for this training step is defined as:
\begin{align}
\vspace{-4mm}
\label{eq:l1}
\mathcal{L}_{R}(\hat{I}^R, I^O) =  ||I^{O}- \hat{I}^{R}||_1
\enspace.
\vspace{-4mm}
\end{align}
From the reconstruction, the network learns to fill the garment texture and locate the desired design to output the original unmasked image $I^O$. We empirically find that this learning step is critical to the full model performance. During the training, we apply random rotations, translations, and scale shifts to the input edge maps. Thus, the model is trained to handle potential geometric misalignment and can locate the desired fashion pattern at the correct position.

\noindent \textbf{Self-Attention Mask Operation (SAM).} \quad During the reconstruction step, the model should learn to change only the target part and keep the rest of an image untouched. Thus, we introduce a self-attention mask to the generator. The self-attention mechanism can guide the model to focus on the target region. Concretely, the decoder/generator $G$ produces two outputs: single-channel self-attention mask $\bf m$ and new pixel image $\bf C$. The final output combines the masked color image with the input cropped image $I^M$:
\begin{equation}
\label{eq:sam}
\hat{I}^R_{i,j}={\bf{m}}_{i,j}\times {\bf{C}}_{i,j} + (1-{\bf{m}}_{i,j}) \times I^M_{i,j} \enspace ,
\end{equation} 
where ${\bf{m}}_{i,j}$, ${\bf{C}}_{i,j}$ and $\hat{I}^{R}_{i,j}$ are the pixels at $i^{th}$ row and $j^{th}$ column in the self-attention mask, the new pixel image, and the final output image. The self-attention mask layer and the color layer share the bottom transpose convolutional blocks in the decoder. The output of the last transpose convolutional layer is fed into two activation layers: a Sigmoid layer with a single-channel output (self-attention mask) and a hyperbolic tangent layer with three-channel output (the new pixel image). The self-attention mask guides the network to focus on the attribute-related region while training the network in a fully self-supervised manner.

\begin{figure*}[t]
    \centering
    \includegraphics[width=0.98\linewidth]{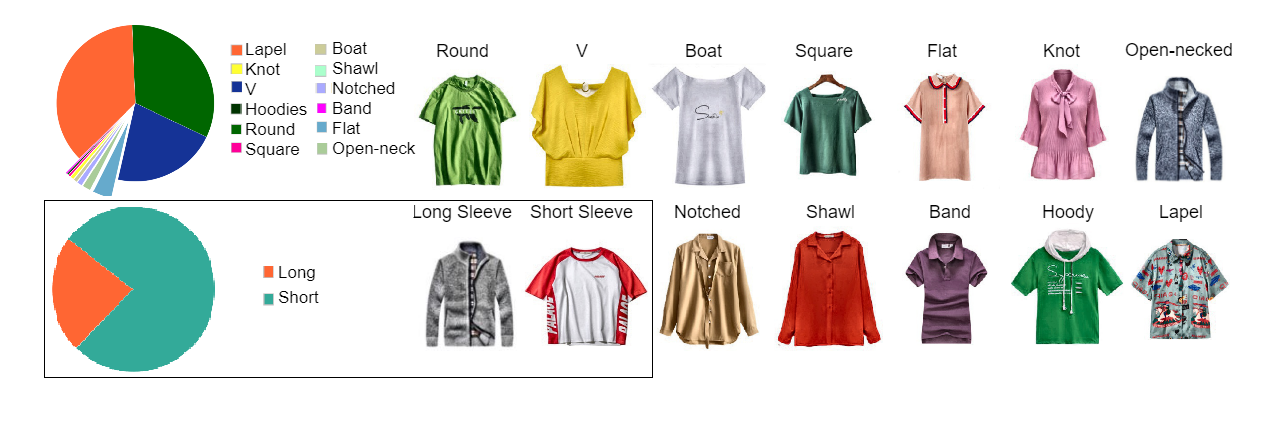}
    \vspace{-4mm}
    \caption{Samples of each collar type and sleeve type from GarmentSet dataset. The pie charts demonstrate the collar type and sleeve type distributions. There are in total 12 collar types and 2 sleeve types.}
    \label{collarSamples}
\end{figure*}

\subsection{Generalized Single Attribute Manipulations}
\label{subsec:gan}

We have introduced the self-supervised reconstruction learning in Sec.~\ref{subsec:encoder-decoder}, which can reconstruct the original image. However, our task is to synthesize new images by manipulating the attributes. The model trained with the reconstruction step could not yield good results since it is not generalized to synthesize a new image with other new attributes (\eg, new collar type, and new sleeve type). Meanwhile, the synthesized image with the reconstruction step is blurry, which makes the results unrealistic. To tackle those problems, we have another adversarial learning step, which inherits the encoder-decoder network from the reconstruction step and applies a novel attribute-aware discriminator.

\noindent \textbf{Perceptual Loss.} \quad To improve the image quality and preserve the overall texture of the original garment image $I^O$, we use VGG perceptual loss:
\begin{align}
\vspace{-4mm}
\label{eq:perceptual}
\mathcal{L}_{\textit{VGG}}(\hat{I}^{T}, I^O) =   ||\phi(I^O)- \phi(\hat{I}^{T})||_1
\enspace,
\vspace{-4mm}
\end{align}
where $\phi$ is a VGG-19~\cite{VGG} feature extractor pre-trained on ImageNet~\cite{imagenet_cvpr09}. 

\noindent \textbf{Attribute-Aware Losses.} \quad To enforce the model to output image with correct attributes, we propose a discriminator with two different regressions: a binary (real/fake) label and an attribute prediction vector. The attribute prediction is optimized by cross-entropy loss, which is defined as:
\begin{align}
\label{eq:class_loss_g}
\mathcal{L}_{ATT}(I, \vec{V})&= -\vec{V} * log(f(I)) \nonumber \\
    &-(1-\vec{V}) * log(1-f(I))
\enspace ,
\end{align}
where $\vec{V}$ denotes the one-hot vector of the target attribute type and ${f}$ is an attribute classifier that outputs the class label vector of image $I$. During the adversarial training, when we forward $I^O$ and its corresponding $ \vec{V}^{O}$ to the discriminator, the discriminator's parameters are updated to learn how to classify the attributes; when we forward $\hat{I}^T$ and the desired target $\vec{V}^T$ to the discriminator, we do not update the discriminator's parameters, but update the generator's parameters. 

By predicting the attribute type, ideally, the discriminator should focus on the replaced design region. Therefore, we introduce an additional feature-level distance measure between $\hat{I}^T$ and $I^T$ to further improve the image quality at the perception-level: 
\begin{align}
\vspace{-4mm}
\label{eq:conceptual}
\mathcal{L}_{\textit{CNT}}(\hat{I}^{T}, I^T) = ||D_{conv}(I^T)- D_{conv}(\hat{I}^{T})||_1,
\vspace{-4mm}
\end{align}
where $D_{conv}$ denotes the feature output from the convolution layers of the discriminator.

By combining  the losses above, the full loss to train the generator can be formulated as:
\begin{align}
\label{eq:gloss}
\mathcal{L}_{\text{G}} &= (1- {\text{D}}( {\text{G}}(E^T,I^M))^2 + \mathcal{L}_{CNT}(\hat{I}^T, I^T)\nonumber \\
 & + \lambda_1 *  \mathcal{L}_{ATT}(\hat{I}^T, \vec{V}^T) +
 \lambda_2 * \mathcal{L}_{VGG}(\hat{I}^T, I^O)
 \enspace ,
\end{align}
where $\lambda_1$ and $\lambda_2$ are hyper-parameters to balance the loss terms. We set $\lambda_1 = 0.1$, $\lambda_2 = 2.5$ in all experiments. In our empirical study, we observe that the model is sensitive to $\lambda_2$ and we need to choose different $\lambda_2$ if we use different VGG layer features to compute the perceptual loss. Similarly, the full loss to train the discriminator is:
\begin{align}
\label{eq:dloss}
\mathcal{L}_{\text{D}} &= \frac{1}{2} [{\text{D}}( {\text{G}}(E^T,I^M))^2 + ({\text{D}}(I^O) - 1)^2] \nonumber \\ 
& + \mathcal{L}_{ATT}(I^O, \vec{V}^O)  
\enspace.
\end{align}

\subsection{Training Algorithm}
\label{subsecL:algo}
\begin{algorithm}[t]
\caption{Training steps}
  \footnotesize
\begin{algorithmic}
\State {\bf Input:}{ learning rate $\alpha$, batch size B }
\State {\bf Output:} { generator parameter $\theta_G$ , discriminator parameter $\theta_D$}
\For {number of iterations}
\State Sample $\{ I^O_i\}_{i=1}^B$ 
\State Extract edge maps and masked-out images $\{E^O_i, I^M_i \}_{i=1}^B$ \
\State $\{\hat{I}^{R}_i\}^B_{i=1}\leftarrow G_{\theta}(\{E_i^O, I^M_i\}^B_{i=1} )$ \
\State \bf{Update the generator G in the reconstruction step:} \
\State $\theta_G \leftarrow Adam\{ \frac{1}{B}\sum_{i=1}^{B} \nabla_{\theta_G} \mathcal{L}_R(\hat{I}_i^{R}, I_i^O),\alpha\}$  
\EndFor
\For {\normalfont number of iterations}
\State Sample $\{ I^O_i, \vec{V}^O_i, I^T_i, \vec{V}^T_i\}_{i=1}^B$
\State Extract edge maps and masked-out images $\{E^T_i, I^M_i \}_{i=1}^B$ \
\State $\{\hat{I}^T_{i}\}^B_{i=1} \leftarrow G_{\theta}(\{E_i^T, I_i^M\}^B_{i=1})$ \
\State \bf{Update the discriminator D in the adversarial step:} \
\State ${\theta_D} \leftarrow Adam\{  \frac{1}{B}\sum_{i=1}^{B}  \nabla_{\theta_D}  \mathcal{L}_{{D}}(\hat{I}_i^{T}, I_i^O, \vec{V}^O_i), \alpha \}$ \
\State \bf{Update the generator G in the adversarial step:} \
\State $\theta_G \leftarrow Adam\{ \frac{1}{B}\sum_{i=1}^{B} \nabla_{\theta_G} \mathcal{L}_{G}(\hat{I}^T_{i}, I^T_i, \vec{V}^T_i, I^O_i), \alpha \}$ 
\EndFor
\end{algorithmic}
\label{al:al1}

\end{algorithm}
Combining the self-supervised reconstruction learning step (Sec.~\ref{subsec:encoder-decoder}) with the adversarial learning step (Sec.~\ref{subsec:gan}), our full model is trained in two separate training loops. Specifically, we formulate the training algorithm in Algorithm~\ref{al:al1}. When training the generator G in the reconstruction step without discriminator D, the generator G receives a masked image $I^M$ and its original edge map $E^O$ as input, and it outputs the reconstructed image $\hat{I}^R$. We try to minimize the reconstruction loss $\mathcal{L}_R$ to enforce the network to learn to generate correct texture based on design sketch and fit to other parts. When training the discriminator D in the adversarial step, the generator G receives masked image $I^M$ and the edge map $E^T$ of a new attribute type as input, and it outputs the edited image $\hat{I}^T$. We try to minimize the generator loss $\mathcal{L}_{G}$, since there is no paired ground truth in this step. It enforces the network to learn to synthesize images by manipulating the target attribute type $E^T$. Then we update parameters in generator G in the adversarial step by minimizing the discriminator loss $\mathcal{L}_{D}$. By optimizing the loss in reconstruction and adversarial steps, the TailorGAN can learn realistic texture and geometry information to yield high-quality images with user-appointed attribute type.
\begin{table*}[t]
\footnotesize

    \centering
  \begin{tabular*}{\linewidth}{  p{1.1cm} p{0.41cm} p{0.44cm} p{0.55cm}|p{0.41cm} p{0.44cm} p{0.55cm} |p{0.41cm} p{0.44cm} p{0.55cm}|p{0.41cm} p{0.44cm} p{0.55cm}|p{0.41cm} p{0.44cm} p{0.55cm}|p{0.41cm} p{0.44cm} p{0.55cm}}
      \toprule
      \toprule
 Type& \multicolumn{3}{c}{Type 1 $\Rightarrow$ Type 2} & \multicolumn{3}{c}{Type 2 $\Rightarrow$ Type 1} & \multicolumn{3}{c}{Type 1 $\Rightarrow$ Type 6} & \multicolumn{3}{c}{Type 6 $\Rightarrow$ Type 1} & \multicolumn{3}{c}{Type 2 $\Rightarrow$ Type 6} & \multicolumn{3}{c}{Type 6 $\Rightarrow$ Type 2}\\
      \midrule 
{}&{C.E.}&{SSIM}&{PSNR} &{C.E.}&{SSIM}&{PSNR} & {C.E.} &{SSIM}&{PSNR}&{C.E.}&{SSIM}&{PSNR}& {C.E.} &{SSIM}&{PSNR} &{C.E.}&{SSIM}&{PSNR}\\
\hline
{CycleGAN} &{12.48}      &{ 0.77}        &{18.72} 
           &\bf{1.74}       &{ 0.64}        &{ 13.97}
           &\bf{5.21}    &0.74           & 23.40
           &\bf{2.97}         &0.78           &18.77
           &6.02    &\bf{0.89}           & 18.89
           &10.02        &0.77           &17.63\\
\hline
{Pix2pix}  &{20.01}      & {0.87}       & {22.75} 
           & 12.73       & \bf{0.89}       & {21.42}
           & 21.62       & 0.88        & {17.63}
           &15.79       & 0.89        & 21.88
           &15.12        & 0.77          &\bf{23.15}
           &19.67        & 0.88          &\bf{23.04}\\
 \hline
   \bottomrule
 {Ours}    &{\bf{9.17}}  &{\bf{0.89}}   &{\bf{23.53}} 
           & 3.70        & \bf{0.89}     & \bf{23.75} 
           & 6.29        & \bf{0.90}     & \bf{23.92}
           &3.25         &\bf{0.90}      &\bf{22.47}
           &{\bf{5.62}}  &\bf{0.89}      & {23.08}
           &\bf{8.13}    & \bf{0.90}     & 22.24\\
      \bottomrule
  \end{tabular*}
 
  \caption{Measurements for all three models based on target translating types on the testing set. C.E. stands for the average classification error score for each paired type translation. The bold numbers in each column are the best scores.}
    \label{tab:main_tb}
\end{table*}

\section{GarmentSet dataset}
\label{sec:dataset}
This part serves as a brief introduction to GarmentSet dataset. Current available datasets like DeepFashion \cite{recomm3} and FashionGen \cite{FashionGen} mostly consist of images including the user's face or body parts and street photos with noisy backgrounds. The redundant information raises unwanted hardness in the training process. To filter out such redundant information in the images, we build a new dataset: GarmentSet. In this dataset, we have 9636 images with collar part annotations and 8616 images with shoulder and sleeve annotations. Both classification types and landmarks are annotated. Although in our studies, the landmark locations are only used in the image pre-processing steps, they still can be useful in future researches like fashion item retrieve, clothing recognition, etc.

In Fig.~\ref{collarSamples}, we present sampled pictures for each collar type, each sleeve type, and the overall data distributions in the dataset. Round collar, V-collar, and lapel images together contribute over eighty percent of the total collar-annotation images. The sleeve dataset only contains two types: short and long sleeves. Although not used in training, the dataset also contains attribute landmark locations, including collars, shoulders, and sleeve ends. 


\section{Experiments}
\label{sec:experiments}

\subsection{Data Pre-processing}
In this paper, we randomly sample $80\%$ data for training and $20\%$ for testing. Meanwhile, in Sec.~\ref{subsec:unseen}, we keep one collar type out in the training set to demonstrate the robustness of our model. In the data pre-processing step, we generate masked-out images $I^M$ and edge maps. We use the left/right collar landmarks to make a bounding box around the collar region. The holistically-nested edge detection (HED) model~\cite{HED} takes charge of making edge maps, which is pre-trained on the BSDS dataset~\cite{MartinFTM01}. 

We notice that the image quality of the edited results depends on the input edge map resolution. Since the HED model used is pre-trained on a general image dataset, it struggles in catching structural details and may also generate unwanted noise maps. The HED model produces low-resolution edge maps for a target image with complicated, detailed structures. 
\subsection{Comparative Studies}
\begin{figure}[t]
    \centering
    \includegraphics[width=0.48\textwidth]{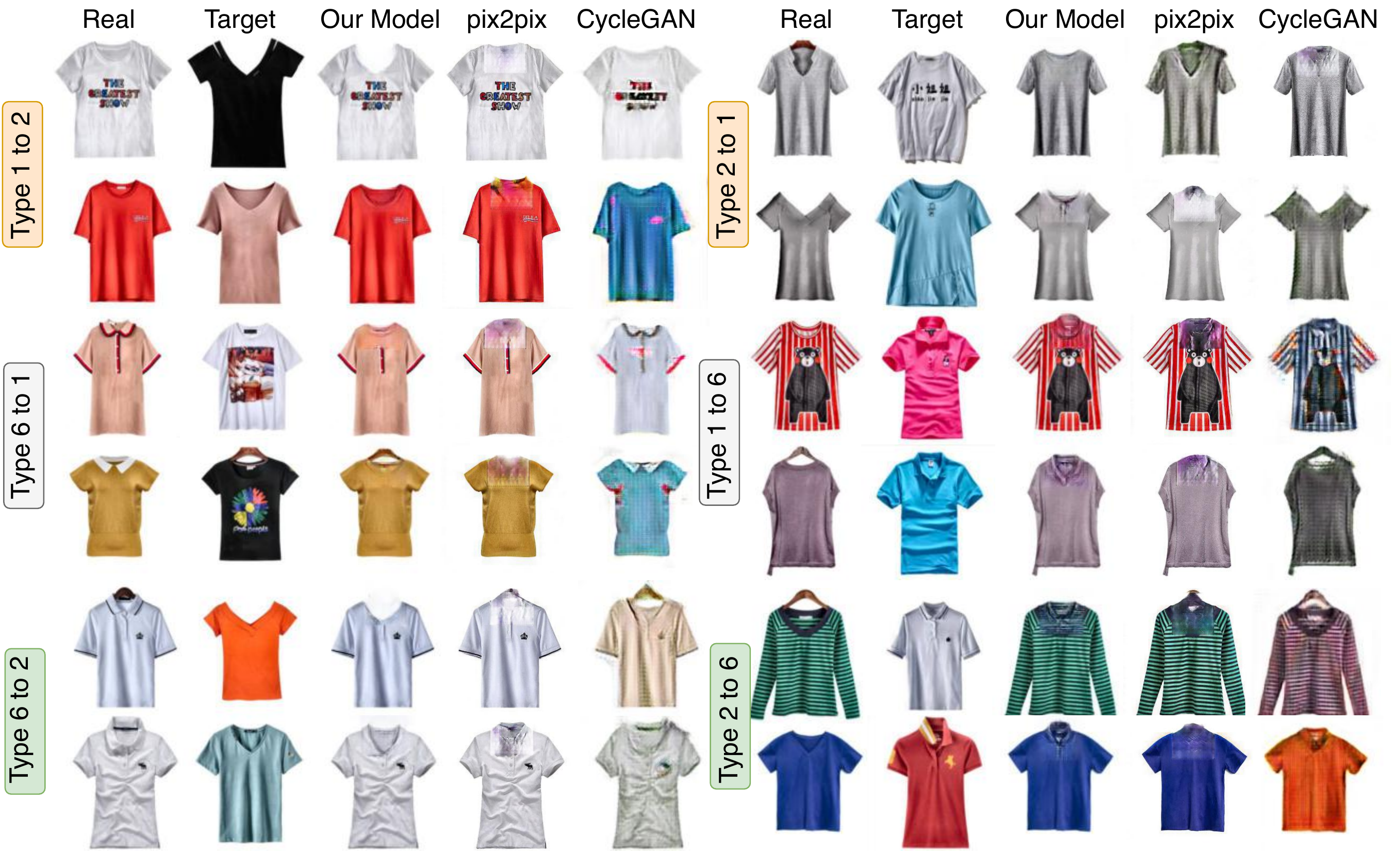}
    \caption{Testing results of collar editing. The images are clustered based on the target collar types. The generated results of simple collar patterns (round and V-collar) are, in general, better than complicated ones (lapel). We train CycleGAN, pix2pix models separately for each specific transformation.}
    \label{fig:collarRes}
\end{figure}
We compare our model with CycleGAN~\cite{cycleGAN} and pix2pix~\cite{p2p}. To compare with \cite{cycleGAN}, we have trained three different CycleGAN models: collar type 1 (round collar) $\Leftrightarrow$ type 2 (V-collar), collar type 1 (round collar)  $\Leftrightarrow$ type 6 (lapel) and collar type 2  $\Leftrightarrow$ type 6. However, our model is trained with all different types together. The results are shown in Tab.~\ref{tab:main_tb}. Furthermore, in Tab.~\ref{tab:randomCompare}, we compare our model with pix2pix using random collar types. Since one CycleGAN model cannot handle translation between any two collar types, we drop it in this random translating comparison test. 

\noindent \textbf{Qualitative results.} \quad In Fig.~\ref{fig:collarRes}, we present testing results of three models. As one can see in the sampled images, CycleGAN does not preserve garment textures. The trained pix2pix model performs poorly in most examples. For the collar part, the pix2pix outputs only show color bulks with no structural patterns. 

\noindent \textbf{Quantitative results.} \quad For the quantitative comparisons, in measuring model performances, we use three metrics: classification errors (C.E.), structure similarity index (SSIM), and peak signal to noise ratio (PSNR). The classification error is measured with a classifier pretrained on GarmentSet dataset. The classification error measures the distance from the target collar designs. The SSIM and the PSNR scores are derived from the differences between the original image and the edited image. From the numerical results presented in Tab.~\ref{tab:main_tb} and Tab.~\ref{tab:randomCompare}, our model outperforms both CycleGAN and pix2pix in making high-quality images with higher classification accuracy. We attribute this to the poor texture/structure preserving of the CycleGAN/pix2pix model.

\begin{table}[t]
\centering
\resizebox{0.6\columnwidth}{!}{
\begin{tabular}{l|l|l|l}
\toprule
\hline
& C.E.    & SSIM            & PSNR           \\ \hline
Our model & \textbf{5.78} & \textbf{0.8921} & \textbf{23.36} \\ \hline
pix2pix   & 16.12         & 0.8783          & 22.83          \\ \hline
\bottomrule
\end{tabular}
}
\caption{Testing results of TailorGAN and pix2pix trained on all collar type inputs. }
\label{tab:randomCompare}
\end{table}

\subsection{Synthesizing Unseen Collar Type}
\label{subsec:unseen}
\begin{figure}[t]
    \centering
    \includegraphics[width=\linewidth]{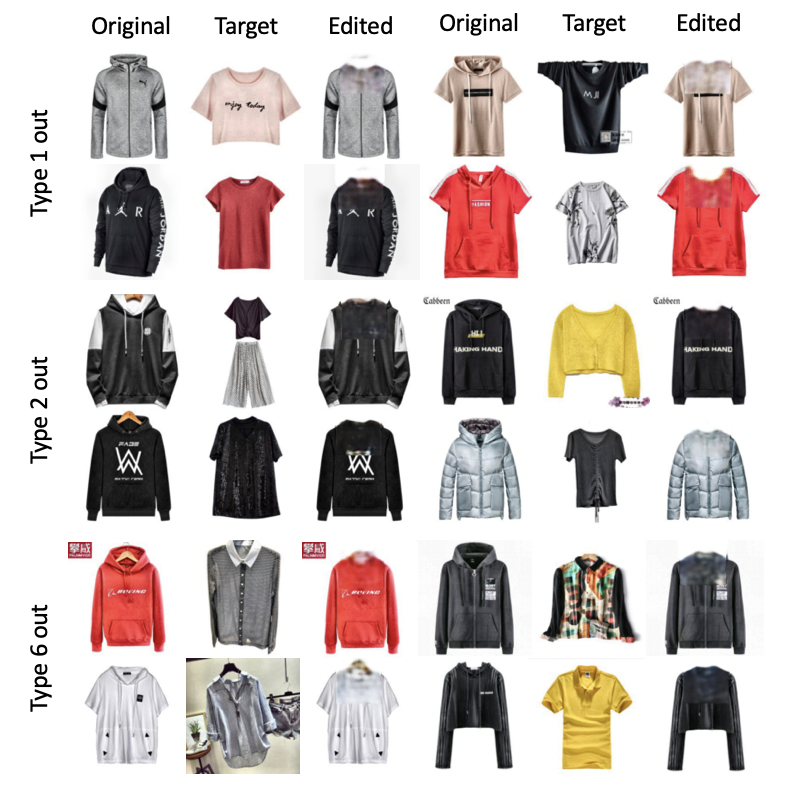}
    \caption{Testing results of the unseen target collar type. The left side indicates which type we are generating (remove this type in training set). The $1^{th}, 4^{th}$ columns are the original reference images. The $2^{th}, 5^{th}$ columns are images with target collar types. The $3^{th}, 6^{th}$ columns are generated images with target collar types with texture/style of reference image.}
    \label{fig:oneout}
\end{figure}

To test our model's capability of processing collar types that are missing in the dataset, we take one collar type out in the training stage and test the model's performance on this unseen collar type. In this test, we take collar type 1, 2, and 6 out. We also test the leave-one-out model with a fully trained model that meets all collar types in its training process. The classification errors for each pair of models are calculated for each taken-out collar type in the testing set. The qualitative results are shown in Fig.~\ref{fig:oneout}

\begin{table}[t]
\centering
\resizebox{0.8\columnwidth}{!}{
\begin{tabular}{l|l|l|l}
\toprule
\hline
C. E.      & type 1 out    & type 2 out   & type 6 out          \\ \hline
full model & \textbf{3.48} & \textbf{7.41} & \textbf{5.27} \\ \hline
one-out model   & 4.74         & 8.59     & 5.34          \\ \hline
\bottomrule
\end{tabular}
}
\caption{The one-out model V.S. the fully trained model.}
\label{tab:oneout}
\end{table}

Based on the qualitative analysis and quantitative comparisons (see Tab.~\ref{tab:oneout}), our model shows a strong generalization ability in synthesizing unseen collar types. 

\subsection{Sleeve Generation}
\begin{figure}[t]
    \centering
    \includegraphics[width=0.95\linewidth]{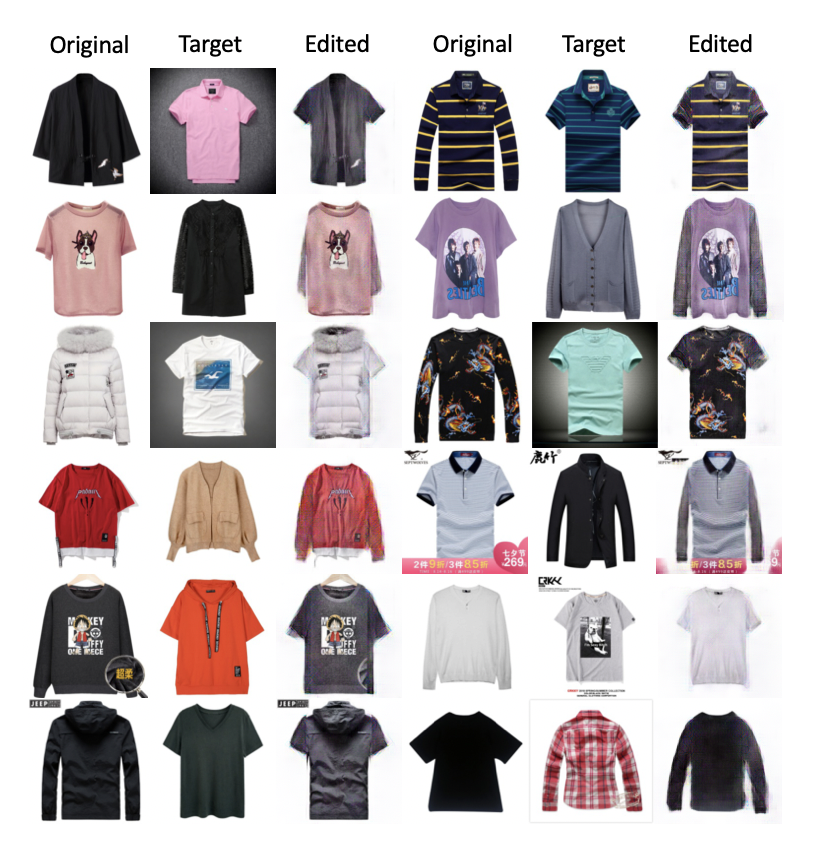}
    \caption{Testing results of editing sleeves using TailorGAN. The $1^{th}, 4^{th}$ columns are the original reference images. The $2^{th}, 5^{th}$ columns are images with target sleeve types. The $3^{th}, 6^{th}$ columns are generated images with target sleeve types with texture/style of reference image.}
    \label{fig:sleeves}
\end{figure}

In the previous discussions, we applied our model in collar part editing. GarmentSet dataset also contains sleeve landmarks and types information. Thus, we test our model's capability in editing sleeves and present testing results in Fig.~\ref{fig:sleeves}. We use the same training scheme. Instead of collars, the sleeve parts are masked out. Due to simpler edge structures and better resolutions in the edge maps, the edited sleeve images have better image qualities and are close to the real images. We discuss more details of sleeve editing results in the user evaluation section. 

\subsection{User Evaluations and Item Retrieves}
\begin{figure}[t]
    \centering
    \includegraphics[width=0.9\linewidth]{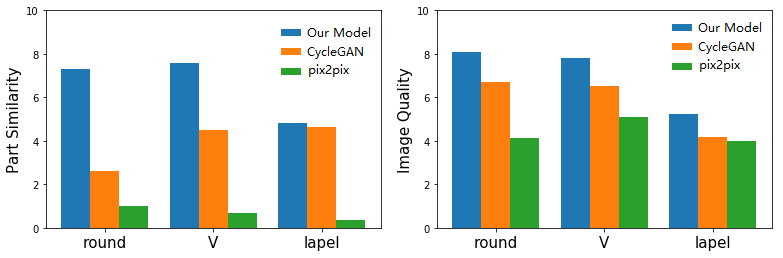}
    \caption{Average user evaluation scores based on image quality and target part similarity. The part similarity is based on users' scores on the edited part structure similarity between the edited image and the target image.}
    \label{fig:userEval}
\end{figure}

To evaluate the performance in a human perceptive level, we conduct thoughtful user studies in this section. Human subjects evaluation (see Fig.~\ref{fig:userEval}) is conducted to investigate the image quality and the attribute (collar) similarity of our generated results compared with \cite{cycleGAN, p2p}. Here, we present the average scores for each model based on twenty users' evaluations. The maximum score is ten. As shown in Fig.~\ref{fig:userEval}, our model receives the best scores in both image quality and similarity evaluations.

\begin{figure}
    \centering
    \includegraphics[width=0.5\linewidth]{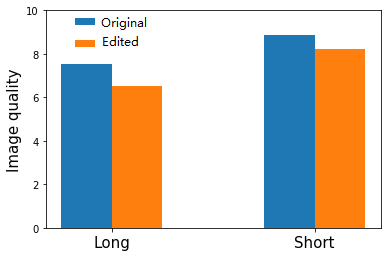}
    \caption{User feedback to sleeve editing results.}
    \label{fig:sleeveEval}
\end{figure}

We also collected users' feedback to the sleeve changing results, and the feedback shows that users can hardly distinguish real/fake between our generated images and the real images. Sleeve tests only evaluate image quality. In each test case, there are two pictures: (1) the original picture and (2) the edited picture. Users decide scores ranging from zero to ten to both pictures based on the image quality. Translating from short to long sleeves is, in general, a harder task due to auto-texture filling and background changing. Users may find that it is harder to distinguish the real images from the edited ones for short sleeve garments. Those observations are reflected in the evaluation scores in Fig.~\ref{fig:sleeveEval}.

\begin{figure}[t]
    \centering
    \includegraphics[width=\linewidth]{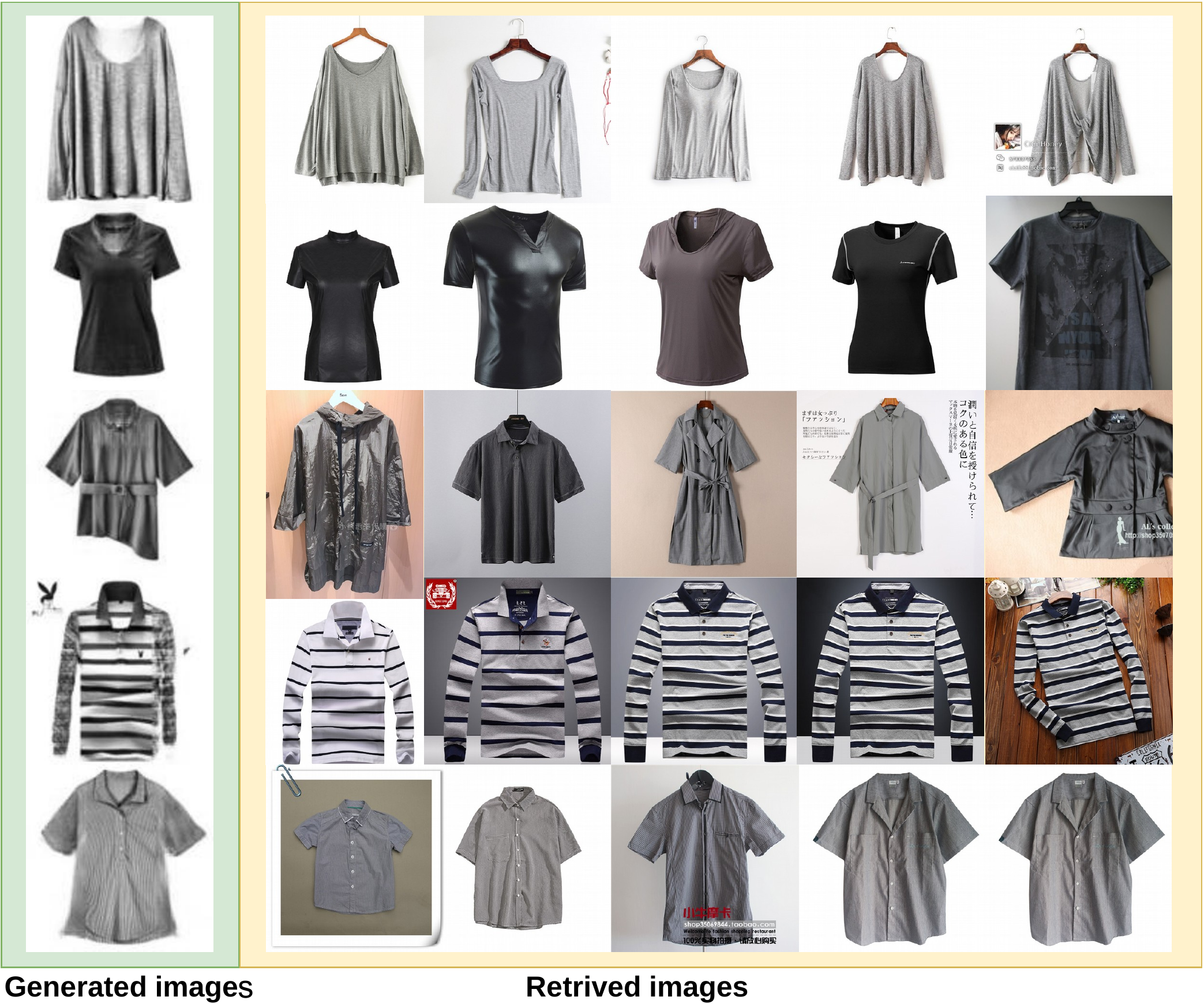}
    \caption{Top 5 matching items from \cite{taobao}. The first column (green box) are generated images and the rest of the column (yellow) are retrieved images based on the generated images from \cite{taobao}.}
    \label{fig:taobao}
\end{figure}

To prove that TailorGAN can be useful in image item retrieves, we upload the edited images to a searching-based website~\cite{taobao} and show top search results in Fig.~\ref{fig:taobao}.

\subsection{Ablation Studies}
\label{sec:ablation}
In this section, we want to clarify our choices of two separate training steps are crucial to generating photo-realistic results. As we argued, to disentangle the texture and the design structure meanwhile keep the rest similar, the model needs to have the ability to reconstruct. Also, using two separate training steps makes sharper results. Tab.~\ref{tabablation} shows that without using the reconstruction training step produces blurry images. Similarly, feeding RGB reference images as input instead of gray-scale edge images, the model produces blurry collars since the lack of clear geometric edge information. For qualitative analysis, We plot testing results of RGB inputs and results of only using adversarial training step and compare those changes with our full model.

\begin{figure}[t]
    \centering
    \includegraphics[width=\linewidth]{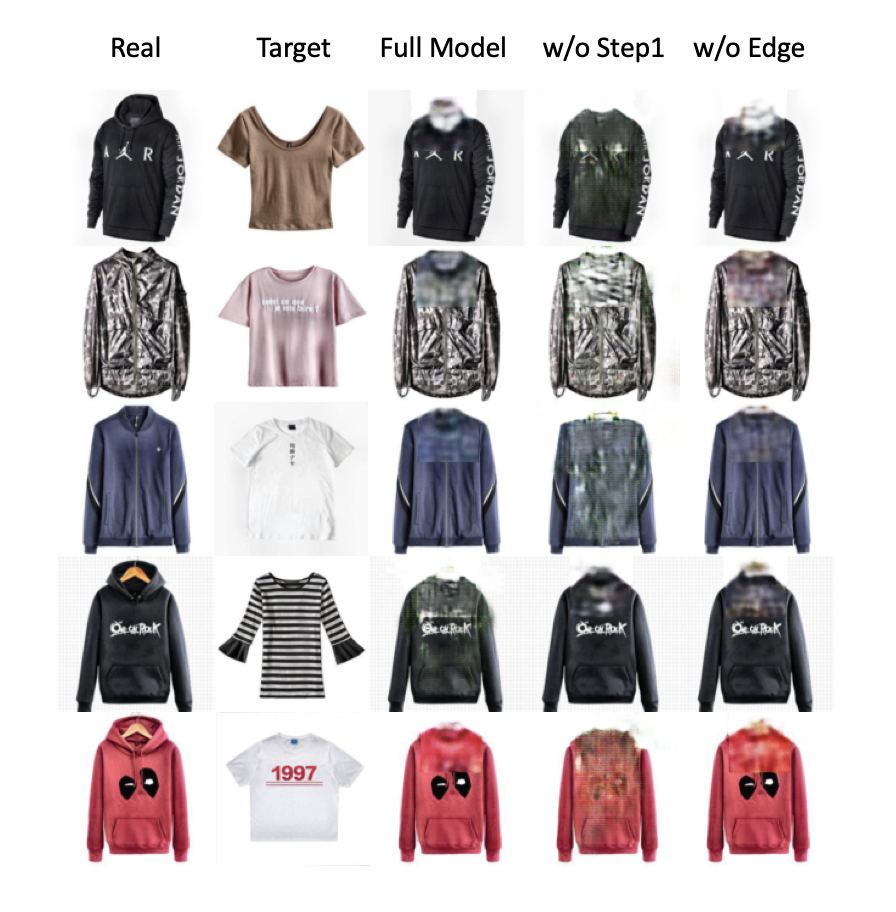}
    \caption{Results of full model, without reconstruction step (w/o reconstruction step) and using RGB pictures (w/o $E^T$) as inputs instead of edge maps.}
    \label{ablation}
\end{figure}

Fig.~\ref{ablation} shows edited image results of our current model versus w/o reconstruction and w/o edge maps. The model w/o reconstruction does not prioritize high-frequency details and tends to average the pixel values in the editing region. On the other hand, using RGB images as inputs, the results show unexpected effects. The RGB-input includes extra structures that are not related to the target images or original images. We hypothesize that those unexpected structures are from the texture-design entanglement. We measure the classification error, SSIM, and PSNR for three variants. Since the major part of the image is left untouched in the result, the leading score may not be impressive in numbers. But, through the qualitative analysis, we can confirm that our current model can generate high-quality images with more details preserved.  

\begin{table}[t]
\centering
\resizebox{0.7\columnwidth}{!}{
\begin{tabular}{l|l|l|l}
\toprule
\hline
              & C.E.    & SSIM            & PSNR           \\ \hline
Full model & \textbf{5.78} & \textbf{0.8921} & \textbf{23.36} \\ \hline
w/o reconstruction step          & 6.34          & 0.7706          & 19.46          \\ \hline
w/o edge input    & 7.21          & 0.8782          & 21.58          \\ \hline
\bottomrule
\end{tabular}
}
\caption{Our final model versus two variants on the testing set. w/o reconstruction step represents we use only adversarial training step rather than two steps. w/o edge input indicates that we use RGB images as input rather than grey-scale edge maps.}
\label{tabablation}
\end{table}

\section{Conclusion}
\label{sec:conclution}
In this paper, we introduce a novel task to the deep learning fashion field. We investigate the problem of doing image editing to a fashion item with user-defined attribute manipulation. Methods developed under this task can be useful in real world applications such as novel garment retrieval and assistant to designers. We propose a novel training schema that can manipulate a single attribute to an arbitrary fashion image. To serve a better model training, we collect our own GarmentSet dataset. Our model outperforms the baseline models and successfully generates photo-realistic images with desired attribute manipulation. \\

\noindent \textbf{Acknowledgement.} \quad This work was partly supported by a research gift from Viscovery. 

{\small
\bibliographystyle{ieee}
\bibliography{egbib}
}

\end{document}